\documentclass{article} 
\usepackage[final]{colm2026_conference}

\usepackage{microtype}
\usepackage{hyperref}
\usepackage{url}
\usepackage{booktabs}
\usepackage{booktabs}
\usepackage{amsmath}
\usepackage{multirow}
\usepackage{macros}
\usepackage{wrapfig}
\usepackage{xcolor}
\usepackage{graphicx}
\usepackage{enumitem}

\usepackage{lineno}

\definecolor{darkblue}{rgb}{0, 0, 0.5}
\hypersetup{colorlinks=true, citecolor=darkblue, linkcolor=darkblue, urlcolor=darkblue}

\title{PubSwap: Public-Data Off-Policy Coordination for Federated RLVR}

\author{Anupam Nayak$^{1*}$, Baris Askin$^{1*}$, Muhammed Ustaomeroglu$^{1}$, Carlee Joe-Wong$^{1}$,\\
\textbf{Gauri Joshi}$^{1}$\\
$^{1}$ Carnegie Mellon University  \\
\texttt{\{anupamn, baskin, mustaome, cjoewong, gaurij\}@andrew.cmu.edu}\\
* Equal contribution
}
\begin{document}

\ifcolmsubmission
\linenumbers
\fi

\maketitle

\begin{abstract}
Reasoning post-training with reinforcement learning from verifiable rewards (RLVR) is typically studied in centralized settings, yet many realistic applications involve decentralized private data distributed across organizations. Federated training is a natural solution, but scaling RLVR in this regime is challenging: full-model synchronization is expensive, and performing many local steps can cause severe client drift under heterogeneous data. We propose a federated RLVR framework that combines LoRA-based local adaptation with public-data-based off-policy steps to improve both communication efficiency and cross-client coordination. In particular, a small shared public dataset is used to periodically exchange and reuse response-level training signals across organizations, providing a lightweight anchor toward a more globally aligned objective without exposing private data. Our method selectively replaces locally incorrect responses with globally correct ones during public-data steps, thereby keeping training closer to the local policy while still benefiting from cross-client coordination. Across mathematical and medical reasoning benchmarks and models, our method consistently improves over standard baselines. Our results highlight a simple and effective recipe for federated reasoning post-training: combining low-rank communication with limited public-data coordination.

\end{abstract}
\section{Introduction}
Reinforcement learning with verifiable rewards (RLVR) has become the central component of post-training for obtaining large language models (LLMs) with strong reasoning ability. Unlike purely supervised approaches such as supervised fine-tuning (SFT) or distillation, RLVR methods like variants of Group relative policy optimization (GRPO) \citep{shao2024deepseekmath} directly optimize models based on outcome-level reward signals derived from verifiable tasks, which has led to substantial gains in reasoning performance. Its success has driven adoption across a wide range of domains with checkable rewards, including mathematics \citep{shao2024deepseekmath, yu2025dapo}, coding \citep{zhu2024deepseek}, scientific reasoning \citep{gunjal2025rubrics,narayanan2025training}, and medical decision-making \citep{gunjal2025rubrics, pan2025medvlm}, finance \citep{zhu2025dianjin, liu2025fin} and lately to even incentivize reasoning via next token prediction \citep{dong2025reinforcement}.

In many important settings, the data required for reasoning post-training is mostly decentralized, with only a limited amount of public data available. This is especially true in personalized \citep{li2026personalized} and domain-specific applications such as medicine, finance, cybersecurity, etc where relevant data is distributed across multiple organizations and cannot be shared freely because of privacy, confidentiality, and leakage concerns. For instance, medical reasoning input prompts may corres to patient histories, diagnosis reports, or clinical notes, while financial reasoning may rely on proprietary transaction data, internal risk reports, or client-specific portfolio information. This makes data sharing and centralized training impractical, while relying solely on isolated private datasets can limit data diversity and reduce generalization beyond each local domain. Federated learning \citep{mcmahan2017communication} offers a natural framework for this setting. Instead of requiring organizations to pool their private data, federated learning trains a shared global model through distributed local steps, allowing data to remain on-device or within each institution. This paradigm has therefore emerged as a promising approach for building models from decentralized data while mitigating privacy and data-sharing concerns.

However, the scale of modern reasoning models makes federated training challenging. Frequent synchronization under full fine-tuning (FFT) incurs substantial communication overhead, as large model updates must be repeatedly transmitted and aggregated across clients. In some settings, gradients or optimizer states may also be maintained at higher precision, further increasing the systems cost. Standard approaches for reducing communication in federated learning include quantization \cite{reisizadeh2020fedpaq, shlezinger2020uveqfed}, sparsification \citep{sattler2019robust, li2023analysis}, and sketching \citep{rothchild2020fetchsgd}. 
While effective in many regimes, these methods generally induce a communication–accuracy tradeoff, and compression can degrade downstream quality especially for reasoning-centric LLMs \citep{liu2025quantization}. This motivates the use of Parameter Efficient Finetuning (PEFT) based methods for Federated RLVR.

A complementary way to amortize communication is to perform more local steps between synchronization rounds. However, under heterogeneous client data, increasing local computation can exacerbate client drift and slow convergence \citep{wang2020tackling}. 
Moreover, most of these methods do not exploit the limited public data often available in decentralized settings, especially for language tasks, where small amounts of public text data are typically accessible.
Even when modest in size, such public data can provide a shared reference for coordination across clients under heterogeneous local distributions, an idea related in spirit to federated distillation \citep{jeong2018communication, lin2020ensemble}.

To address these challenges, we propose a two-fold solution in the form of a federated post-training framework.
In our approach, each client performs local GRPO steps using Low-Rank Adaptation (LoRA) \citep{hu2022lora} rather than full fine-tuning, reducing both communication cost and client-side memory usage. In addition, we leverage shared public data to perform off-policy updates that provide lightweight synchronization across clients during local GRPO training.

In particular, beyond standard federated parameter exchange, our method communicates evaluations of \textit{public} data points produced by models hosted on local clients, which do not introduce additional privacy concerns.
Together, these two components enable communication-efficient federated RLVR while preserving the ability to benefit from both private local data and limited shared public data. The remainder of the paper is organized as follows.
\begin{itemize}
    \item Section~\ref{sec:related works} reviews the prior work relevant to our setting, and Section~\ref{sec:propmeth} introduces the problem formulation and our proposed methods.
    \item Section~\ref{sec:expres} presents empirical results demonstrating the effectiveness of our proposed $\dist$ method with $\keep$ response aggregation on both mathematical and medical reasoning tasks on Qwen and LLama family of models, along with ablations on key hyperparameters. Section~\ref{sec:conc} concludes the paper.
    \item Additional experimental details, as well as theoretical derivations upper bounding the per-step drift induced by both our method and the FedAvg baseline, are deferred to the appendix.
\end{itemize}
\section{Related Works}
\label{sec:related works}
This work was inspired by the following previous studies conducted in related areas.

\textbf{Decentralized training of Reasoning Models:} Recent work has begun to examine distributed training paradigms for large models. In the large scale private-data setting, prior studies have considered collaborative training without direct data sharing, for example through localized mixture-of-experts (MoE) training followed by model merging \citep{shi2025flexolmo}. MoE parameter efficient finetuning (PEFT) have also been explored for on-device training \citep{fan2025device, wagnerpersonalized, singhal2026fedsb, raje2025ravan}. Meanwhile, \citet{zhang2024towards} show that FedAvg-style optimization can be effective for instruction tuning. However, these works are largely restricted to supervised training with next-token prediction losss. In the RLVR setting, recent works study decentralized training \citep{team2025intellect, zhang2026gepo}. While they face related challenges, including communication overhead and drift arising from stale or off-policy updates, they focus on distributed training over shared public data rather than private data distributed across clients. Another closely connected application domain is personalized reasoning \citep{li2026personalized, li2025prefpalette}, In a federated formulation, \citet{wang2026fedmoa} study GRPO based LLM training for personalized reasoning under heterogeneous rewards. Recent work has also extended federated learning to reinforcement-learning-based training of LLM agents, including GRPO-style methods \citep{chen2026federated}.

\textbf{Learning from public and private data:} The use of representative public data has long been studied in federated learning. A common application is federated distillation, where public inputs are passed through client models trained on private data, and the resulting logits are used as teacher signals to distill a server-side model \citep{jeong2018communication, lin2020ensemble, cho2022heterogeneous, shao2024selective}. In LLM settings, public data has been used in several ways. The most common approach first tunes on public data and then fine-tunes or post-trains on private data for personalization or downstream adaptation \citep{yu2024privacy, hanke2024open}. More recent work studies combining public and private data across multiple training stages, including during pretraining \citep{bu2024pre} and through public mid-training followed by federated learning on private data \citep{wang2024can}. Another line of work uses public data to train generative models that synthesize samples matching private-data distributions \citep{yu2024privacy, hou2025private}. More broadly, the value of public data has also been studied extensively in differential privacy \citep{alon2019limits}; most closely related to our setting, \citet{jiang2025cost} analyze a public-private interleaving scheme for gradient-based optimization similar to ours.

\textbf{Off policy Learning and GRPO}: GRPO has been a key driver of recent progress in reasoning language models, motivating substantial effort toward improving its efficiency in terms of memory, wall-clock time, and compute. A recurring theme in this literature is off-policy training, where rollouts generated by a different behavioral policy, stale or otherwise non-current policies are reused to accelerate learning. Asynchronous GRPO based distributed training \citep{team2025intellect, zhang2026gepo} is one important special case of this broader setting, since asynchrony naturally leads to stale-policy updates. Because such reuse introduces policy mismatch, it can degrade optimization through bias, instability, and drift. Previous work mitigates these effects through techniques such as careful importance sampling, reward shaping, asymmetric baselines, and clipped surrogate objectives \citep{zheng2026prosperity, wan2026buffer, mroueh2025revisiting, yan2025learning, arnal2025asymmetric}. These techniques are complementary to our approach and could be incorporated on top of our method to further improve performance.

\textbf{Efficient Federated Training:} Federated learning (FL) is the dominant paradigm for distributed training without direct data sharing. A large body of work has focused on improving FL by accelerating convergence \citep{jhunjhunwalafedexp, pathak2020fedsplit}, reducing client drift under data heterogeneity \citep{li2020federated, wang2020tackling, karimireddy2020scaffold, acar2021federated}, and improving communication efficiency \citep{konevcny2016federated, reisizadeh2020fedpaq, shlezinger2020uveqfed, pmlr-v238-zakerinia24a}. However, these methods were largely developed for smaller-scale models and do not directly address the optimization, memory, and systems challenges that arise in LLM training. In the LLM setting, a common approach to communication- and memory-efficient federated adaptation is to combine FL with parameter-efficient fine-tuning methods such as LoRA \citep{wagnerpersonalized, singhal2026fedsb, raje2025ravan, cho2022heterogeneous, zhang2024towards}. Although effective, these methods primarily focus on supervised fine-tuning objectives and are less suited beyond these settings to domains like RL post training.

\section{Proposed Method}
\label{sec:propmeth}
\subsection{Preliminaries}
\textbf{GRPO:} Group Relative Policy Optimization (GRPO) \citep{shao2024deepseekmath} is a policy gradient method widely used for RL post training of LLMs. It replaces the critic model used to compute the advantage function in the Proximal Policy Optimization (PPO) \citep{schulman2017proximal} with a relative advantage of the group. Let $(y_1\cdots y_\K) \sim \pi(\cdot|x)$ be IID responses produced by the model $\pi$ to a prompt $x$ then the advantage is computed as
\begin{align}
    A_k(x,y_k) :=  \frac{r(x,y_k) - \bar{r}(x)}{\sigma(x)} \qquad\forall k\in \{1,2\cdots \K\},
\end{align}
where $r(x,y)$ is the reward associated with the response $y$ for the prompt $x$, $\bar{r}(x)$ and $\sigma(x)$ are the group-mean and standard deviation across the responses $\{y_1\cdots y_K\}$ to the same prompt $x$.
$\{y_1\cdots y_\K\}\sim\pi_{\tht_{\text{old}}} (\cdot|x)$ are generated using a policy parameterized by $\tht_{\text{old}}\in \Tht$. 
The GRPO objective is then given as
\begin{equation}
J(\theta)
=
\mathop{\mathbb{E}}\limits_{x \sim \mathcal{D}}
\left[
\frac{1}{K}
\sum_{k=1}^{K}
\frac{1}{|y_k|}
\sum_{t=1}^{|y_k|}
\min \!\left(
\frac{\pi_{\theta}(y_{k,<t}|x)}{\pi_{\theta_{\mathrm{old}}}(y_{k,<t}|x)} A_k,
\operatorname{clip}\left(
\frac{\pi_{\theta}(y_{k,<t}|x)}{\pi_{\theta_{\mathrm{old}}}(y_{k,<t}|x)},
 1-\epsilon, 1+\epsilon
\right) A_k
\right)
\right]
\label{grpo_obj}
\end{equation}
where $A_k$ is the shorthand for $ A_k(x,y_k)$ and $y_{k<t}$ denotes all tokens in $y_k$ before position $t$. A GRPO step has two substeps: (1) generate responses for the minibatch of prompts using $\pi_{\tht_{\text{old}}}$, which defines the objective, and (2) update the policy via gradient ascent, possibly with multiple ascent iterations (GRPO update). The updated policy then replaces $\pi_{\tht_{\text{old}}}$ for the next GRPO step.

\subsection{Setup}
We consider the standard federated learning framework consisting of $N$ clients and a central server, widely used in previous literature \citep{jeong2018communication,mcmahan2017communication}. The local dataset at each client $n \in [N]$ consists of query-verifiable answer pairs and is denoted by $D_n := \{(x_i^{(n)}, y_i^{(n)})\}_{i=1}^{|D_n|}$.
In addition, there exists a shared public dataset $D_{\text{pub}}$ that serves as a representative reference set, is accessible to all clients and the server.  The public dataset is typically much smaller than the total amount of private training data residing across clients. For example, in a cross-hospital medical imaging setting, hospitals may retain private MRI datasets locally, while a public MRI dataset serves as a common reference set for everyone.
The goal is to train a shared model that performs well across the query distributions of all clients, as measured by its performance on a held-out test dataset maintained by the server. This test data set is typically assumed to be representative of global training data, which is distributed heterogeneously across clients enabling the evaluation of the extent to which client-specific knowledge contributes to the learned model.

\subsection{LoRA for communication-efficient federated aggregation}

All clients are initialized from a shared pretrained model at the beginning of training. To reduce both the memory footprint of local fine-tuning and the communication overhead incurred during synchronization, we parameterize client updates using Low-Rank Adaptation (LoRA) \citep{hu2022lora}. Recent work suggests that LoRA is especially effective for RLVR \citep{wang2025tina}, possibly because the RL training use the reward signal contains much less information than token-level supervised fine-tuning labels and can therefore often be captured and encoded easily using low-rank updates \citep{schulman2025lora}. 

Specifically, for a layer weight matrix in the pretrained backbone, we parameterize the global model at round $\rn$ as
\begin{equation}
\thl_{\rn} = \thl_{0} + \B_{\rn}\A_{\rn},
\end{equation}
where $\thl_0 \in \mathbb{R}^{m \times d}$ denotes the frozen pretrained weight matrix, $\B_{\rn} \in \mathbb{R}^{m \times r}$ and $\A_{\rn} \in \mathbb{R}^{r \times d}$ are the global LoRA factors at round $\rn$, and $r \ll \min\{m,d\}$ is the adaptation rank. At the beginning of each round, the server broadcasts $\left(\A_{\rn}, \B_{\rn}\right)$ to all clients. Client $n$ initializes its local factors from the global ones and, after local training, obtains updated factors $\left(\A_{\rn+1}^{(n)}, \B_{\rn+1}^{(n)}\right)$.
During local training, the backbone $\thl_0$ remains fixed and only the LoRA factors are optimized. We initialize $\B_{0}$ to the zero matrix and $\A_{0}$ randomly before training starts.

To preserve the communication benefits of low-rank updates during federated synchronization, we adopt the inexact aggregation strategy of FedIT \citep{zhang2024towards}. Rather than transmitting full dense parameter updates, each client communicates only its LoRA matrices, which are averaged separately across clients at the server to obtain the next round's parameters:
\begin{equation}
\B_{\rn+1} = \frac{1}{N}\sum_{n=1}^N \B_{\rn+1}^{(n)},
\qquad
\A_{\rn+1} = \frac{1}{N}\sum_{n=1}^N \A_{\rn+1}^{(n)}.
\end{equation}

This reduces the per-layer communication cost from $2md$ to $2r(m+d)$, which is significantly smaller since $r \ll \min\{m,d\}$.
\subsection{Local Optimization Algorithms}
Here we will list two local optimization algorithms we use in our experiments

\begin{figure}
    \centering
    \includegraphics[width=1\linewidth]{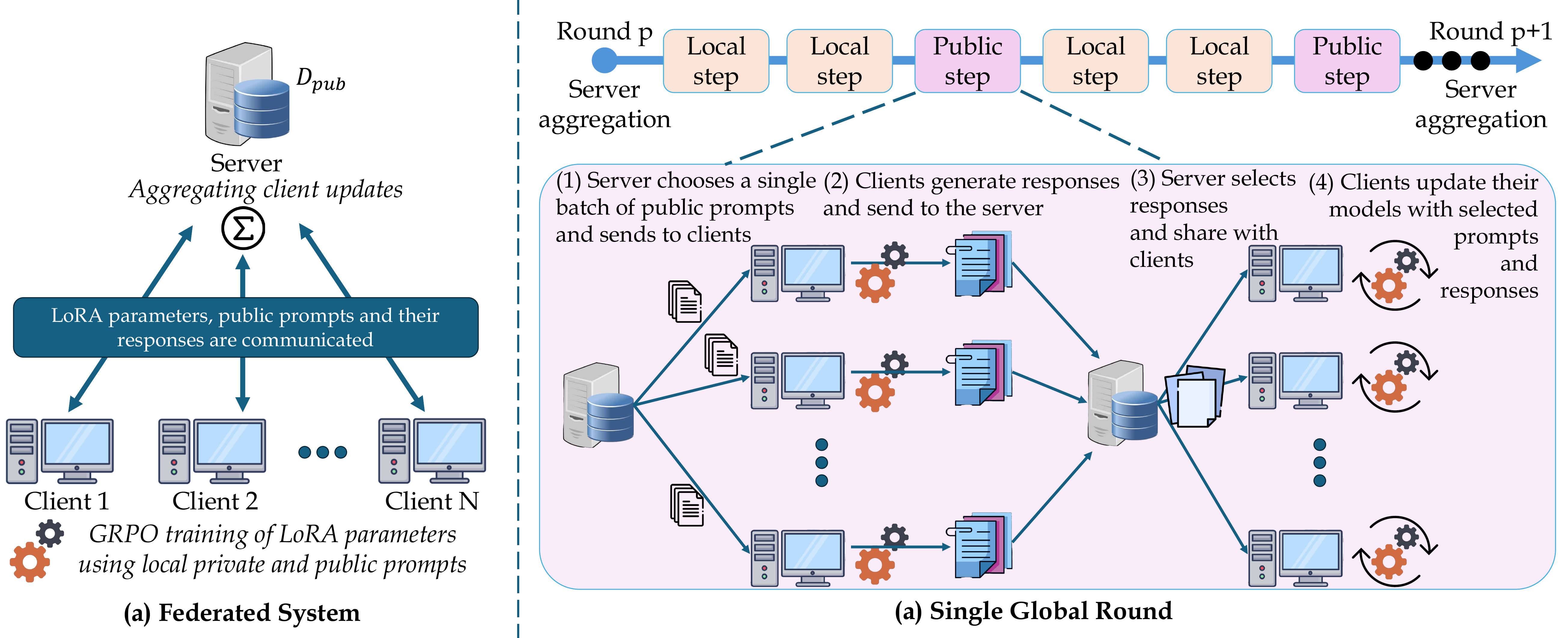}
    \caption{The figure illustrates our proposed $\dist$ method, which alternates training on both public and private data. During a local step, each client performs a GRPO step on minibatches sampled from its own private dataset. During the public step, every client generates $\K$ responses for the same shared batch of prompts and sends these generations to the server. Based on the response aggregation method, the server then returns the responses to each client, which are used them to perform GRPO update. Models are aggregated after each communication round. }
    \label{fig:placeholder}
\end{figure}

\noindent \textbf{Local GRPO:} During each communication round, every client performs $\tau$ local GRPO steps. At every such step, the client samples a size $b$ minibatch of prompts from its local dataset, generates $\K$ responses per sample and applies a fixed number of gradient ascent steps to optimize the GRPO objective (Eq.~\ref{grpo_obj}). Once $\tau$ local steps are completed, the client sends its learned LoRA parameters to the server for aggregation, after which the global parameters aggregated across clients are communicated back to all clients.

\noindent \textbf{\dist:} The primary goal of this procedure is to leverage public data to mitigate client drift while still learning from the broader set of private local data. To this end, we interleave steps on private and public data during local training. In addition to the local-step parameter $\tau$, the algorithm takes as input a $\dist$ period $\tau_{\swap}<\tau$. Within each $\dist$ period, the first $\tau_{\swap}-1$ steps are standard GRPO steps using private data, while the final step is a public step performed locally on shared public data. 

The public-data-based steps here are intended to mitigate client drift. We consider the following strategies for incorporating the public dataset:
\begin{itemize}[leftmargin=*]
    \item \textbf{$\rand$:} The server first selects $\bt$ prompts from the public dataset 
    and communicates this set to all clients. Each client then generates $\K$ answers for each selected prompt and sends them back to the server. From the resulting pool of $\N\K$ answers per prompt, the server uniformly samples $\K$ answers at random for each prompt and broadcasts them to all clients. Each client then performs a GRPO update using this shared global batch of $\K$ answers.
\item \textbf{$\keep$:} 
 Similarly to $\rand$, the server begins by sampling a batch of $\bt$ prompts from the public dataset and broadcasting them to all clients. For each prompt, every client generates $\K$ i.i.d.\ responses and transmits them to the server. Let $c_n$ denote the number of correct responses obtained by client $n$ for a particular prompt. The response pool used for the local GRPO update at client $n$ is constructed according to the following rule:
\begin{itemize}
    \item If $c_n \geq \K/2$, the client forms its GRPO objective using only its own generated responses.
    \item If $c_n < \K/2$, the client replaces up to $\K/2 - c_n$ responses with correct responses drawn from the globally collected pool, excluding its own generations. 
\end{itemize}
This response aggregation strategy increases reward variance within the response set assigned to each client while replacing only locally incorrect responses with globally correct ones.
\end{itemize}

Both proposed methods modify only the first part of the GRPO step, namely response generation and objective definition, by swapping responses and defining the objective accordingly. The communication overhead associated with the public data steps is negligible compared to the cost of exchanging the LoRA matrices at the end of every communication round. 

The proposed mechanisms are inherently off-policy, because part of the response set used for a local GRPO update may come from client models that have already drifted during the preceding private-data steps (See figure \ref{fig:placeholder}). In this sense, the resulting learning dynamics are intuitively closer to SFT, in which models learn from trajectories produced by other models. Moreover, under $\keep$, a local answer is replaced by a global correct answer only when the local answer is incorrect. Consequently, the local policy learns both by downweighting locally incorrect answers and by leveraging both locally correct and globally correct answers to accelerate learning. 

Compared to $\rand$, the $\keep$ strategy is less off-policy in nature. This is because it replaces only locally incorrect responses, and only when the number of locally correct responses falls below $\K/2$. As a result, the local GRPO objective is constructed using a response set that remains closer to the client’s own policy. Moreover, when all responses to a prompt are either correct or incorrect, the GRPO advantage vanishes and the objective in Equation~\eqref{grpo_obj} becomes zero. For this reason, in $\keep$ we replace responses only up to the point where the resulting set contains an equal number of correct and incorrect answers. This maximizes reward variance within the subset \citep{xu2025not}, which has been observed to improve performance \citep{xu2025not,ye2025beyond}.

\section{Experimental Results}
\label{sec:expres}
\begin{table}[t]
\centering
\small
\begin{tabular}{@{}ll|cccc|cccc@{}}
\toprule
\multirow{2}{*}{Model} & \multirow{2}{*}{Method} 
& \multicolumn{4}{c|}{MATH} 
& \multicolumn{4}{c}{DeepMath} \\
& & $\tau$=10 & $\tau$=40 & $\tau$=90 & $\tau$=120 
& $\tau$=10 & $\tau$=40 & $\tau$=90 & $\tau$=120 \\
\midrule

\multirow{4}{*}{Qwen3-1.7B}
& Base model  & 55.2 & 55.2 & 55.2 & 55.2 & 14.9 & 14.9 & 14.9 & 14.9 \\
& FedAvg-GRPO & \textbf{77.9} & 75.6 & 75.5 & 75.9 & 48.9 & 52.7 & 50.4 & 50.7 \\
& FedProx-GRPO & 77.5 & 76.0 & 76.5 & 75.6 & 50.6 & 52.3 & 48.0 & 47.7 \\
& FedAvg-$\dist$ & 77.0 & \textbf{76.7} & \textbf{76.9} & \textbf{76.6} & \textbf{51.1} & \textbf{53.0} & \textbf{53.3} & \textbf{55.8} \\

\midrule

\multirow{4}{*}{\shortstack{Qwen2.5-\\Math-1.5B}}
& Base & 58.4 & 58.4 & 58.4 & 58.4 & 34.9 & 34.9 & 34.9 & 34.9 \\
& FedAvg-GRPO & \textbf{73.5} & \textbf{72.3} & 71.9 & 71.1 & 51.0 & 52.3 & 50.5 & 49.3 \\
& FedProx-GRPO & 72.8 & 71.4 & 71.5 & 70.6 & 52.2 & 52.1 & 48.0 & 49.4 \\
& FedAvg-$\dist$ & 72.4 & 71.6 & \textbf{72.5} & \textbf{71.9} & \textbf{52.4} & \textbf{53.1} & \textbf{50.8} & \textbf{53.1} \\

\bottomrule
\end{tabular}
\caption{Pass @1 performance comparison on MATH \citep{hendrycks2021measuring} and DeepMath \citep{he2025deepmath} across different numbers of local steps ($\tau$). The $\keep$ method and a $\swap$ period of 2 is used here for response aggregation with $\dist$. We tune baselines to the best performance, more details are presented in the Appendix.}
\label{tab:math_deepmath_results}
\end{table}

\subsection{Setup}

\noindent \textbf{Datasets and Models:} We choose the Qwen family of models for our models, presenting results on Qwen2.5-MATH-1.5B \citep{yang2024qwen2}, Qwen3-1.7B and Qwen3-4B-Instruct models \citep{yang2025qwen3}. For math reasoning, we evaluate our method against baselines on MATH \citep{hendrycks2021measuring} and DeepMath \citep{he2025deepmath}. 
MATH contains 12.5K competition-style high-school math problems with difficulty levels 1--5, whereas DeepMath is a much larger 103K-problem benchmark with predominantly higher-difficulty problems, making it substantially harder than MATH. We also test our proposed method on medical reasoning tasks for which we use a mixture of MedQA \citep{jin2021disease} and MedMCQA datasets \citep{pal2022medmcqa} and the Llama3.2-3B-Instruct \citep{grattafiori2024llama} model.

\textbf{Implementation details:} Our implementation is based on the VERL framework \citep{sheng2025hybridflow}, and we adopt most of the recommended hyperparameter settings from VERL. In addition, we apply LoRA to all layers across models, using LoRA rank = 32 and LoRA $\alpha=64$. Consequently, we use a learning rate of $1\times10^{-5}$ rather than the standard $1\times10^{-6}$ recommended in VERL for full fine-tuning, following scaling law evidence suggesting that LoRA typically benefits from a learning rate roughly 10$\times$ larger than that used for full fine-tuning \citep{schulman2025lora}.We additionally adopt the clip higher strategy for importance weights proposed in DAPO \citep{yu2025dapo}, resulting in an asymmetric trust region of $(1-\epsilon_{\text{low}},1+\epsilon_{\text{high}})$, where $\epsilon_{\text{low}}=0.2$ and $\epsilon_{\text{high}}=0.25$. These settings are used consistently across all methods and baselines. Additional details are provided in the appendix.

\begin{wraptable}{h}{0.56\textwidth}
\vspace{-0.8em}
\centering
\small
\begin{tabular}{@{}l|cccc@{}}
\toprule
Method & $\tau$=10 & $\tau$=40 & $\tau$=90 & $\tau$=120 \\
\midrule
Base model       & 51.1 & 51.1 & 51.1 & 51.1 \\
FedAvg-GRPO  & 66.9 & \textbf{66.5} & 67.0 & 67.4 \\
FedAvg-$\dist$      & \textbf{70.0} & 65.4 & \textbf{70.2} & \textbf{70.5} \\
\bottomrule
\end{tabular}
\caption{Pass@1 performance of the model Qwen3-4B-Instruct-2507 on DeepMath across different numbers of local steps ($\tau$). The $\keep$ method and a $\swap$ period of 2 is used here for response aggregation with $\dist$.}
\label{tab:qwen3_4b_deepmath}
\vspace{-1em}
\end{wraptable}

\textbf{Baselines:} We report both the final checkpoint test accuracy and, in the appendix, the best checkpoint accuracy for all runs. In all math reasoning experiments reported in the main text, we train for 360 GRPO steps, with two gradient update steps per GRPO step, and evaluate multiple choices of the number of local steps ($\tau$). As baselines, we compare our method against FedAvg + GRPO and FedProx \citep{li2020federated}, a standard method for mitigating client drift in supervised federated learning. FedProx modifies the local minimization objective on client $k$ to
\begin{equation}
\min_{\theta} \; F_k(\theta) + \frac{\mu}{2}\left\|\theta-\theta^{(t)}\right\|^2,
\end{equation}
where $\theta^{(t)}$ denotes the global model parameters at the start of the communication round, before local training begins. For each reported model, we tuned $\mu \in \{0.001, 0.01, 0.1\}$ separately and report the best-performing choice. In our implementation, the FedProx penalty is applied only to the LoRA matrices separately. For all runs, we use a rollout temperature of 0.7, a maximum generation length of 2048 tokens, and no top-$p$ or top-$K$ sampling. Additional details are provided in the appendix.

\subsection{Main Results}
\textbf{Mathematical reasoning tasks:} Our main results for Qwen2.5-Math-1.5B and Qwen3-1.7B are reported in Table~\ref{tab:math_deepmath_results}. Across both model sizes, our proposed $\dist$ method with $\keep$ response aggregation outperforms both FedAvg+GRPO and FedProx for most choices of the local step parameter. Unless otherwise noted, all reported results for the $\keep$ variant use a $\dist$ period of 2. Notably, the gains are larger at higher values of the local step ($\tau$) parameter. 

This regime is particularly important in practice, since larger $\tau$ reduces synchronization frequency and hence communication overhead. The stronger performance of our method in this setting suggests that it is especially effective in the communication-efficient regime where standard federated RL training becomes more challenging. 

Our experiments further show that, unlike in many supervised federated learning settings,
FedProx does not improve performance in our RLVR setup and in many cases underperforms even vanilla FedAvg+GRPO. One possible explanation is that constraining local updates in parameter space does not necessarily induce a corresponding constraint in policy/token space: for autoregressive models, small differences in token probabilities can compound through sampling and lead to substantially different output trajectories. 

We additionally report results for Qwen3-4B-Instruct model on the DeepMath dataset in Table~\ref{tab:qwen3_4b_deepmath}. Here, $\dist$ outperforms FedAvg+GRPO at three out of four local-step settings and is broadly consistent with our main results.

\begin{wraptable}{h}{0.56\textwidth}
\vspace{-0.8em}
\centering
\small
\begin{tabular}{@{}l|cccc@{}}
\toprule
Method & $\tau$=10 & $\tau$=40 & $\tau$=90 & $\tau$=120 \\
\midrule
Base model      & 49.2 & 49.2 & 49.2 & 49.2 \\
FedAvg-GRPO  & \textbf{58.7} & 58.2 & 57.9 & 56.0 \\
FedAvg-$\dist$      & 57.5 & \textbf{59.4} & \textbf{58.5} & \textbf{58.1} \\
\bottomrule
\end{tabular}
\caption{Pass@1 performance of the model Llama3.2-3B-Instruct on medical reasoning across different numbers of local steps ($\tau$). The $\keep$ method and a $\swap$ period of 2 is used here for response aggregation with $\dist$.}
\label{tab:llama_med}
\vspace{-1em}
\end{wraptable}

\textbf{Medical reasoning tasks:}
Our main results on the medical reasoning tasks are presented in Table~\ref{tab:llama_med}. All the reported numbers correspond to 540 GRPO steps except for the $\tau$=120 setting where the reported accuracy corresponds to 600 GRPO steps. Similar to the mathematical reasoning experiments, our proposed $\dist$ method outperforms the FedAvg+GRPO baseline in the settings where it is evaluated.

A common observation across both setups is that increasing the local-step parameter does not always degrade performance. One possible explanation is that when synchronization is too frequent, aggregation is performed after only a few steps, which can interrupt the optimization dynamics by resetting the local optimizer state at the end of each round. While this effect is less pronounced in smaller-scale settings, it may become more significant for larger models. Prior work has proposed alleviating this issue through server-side momentum \citep{reddi2020adaptive,hsu2019measuring}, and such techniques could in principle be combined with our method to improve performance in the small-local-step regime. However, we expect them to be less important in the more practical setting where the number of local steps is sufficiently large. In our implementation, optimizer states are reinitialized after each aggregation round, so this issue affects both the baselines and the proposed method.

\begin{wraptable}{h}{0.55\textwidth}
\vspace{-0.8em}
\centering
\small
\begin{tabular}{@{}l|ccc@{}}
\toprule
Method & $\tau$=40 & $\tau$=90 & $\tau$=120 \\
\midrule
Base model         & 14.9 & 14.9 & 14.9 \\
FedAvg+GRPO   & 52.0 & 51.6 & 51.3 \\
$\dist$       & \textbf{53.0} & \textbf{56.7} & \textbf{54.9} \\
\bottomrule
\end{tabular}
\caption{Pass@1 performance of the model Qwen3-1.7B on DeepMath across different numbers of local steps ($\tau$) for a higher heterogeneity split. }
\label{tab:qwen3highet}
\vspace{-1em}
\end{wraptable}
Additionally, we find that under the $\dist$ method, $\rand$ can outperform $\keep$ response aggregation scheme when the number of local steps is small (10 or 40). This advantage does not persist at larger local-step values (90 or 120), where training can become unstable, especially for Qwen3-1.7B. One possible explanation is that as client models drift farther apart, responses generated by other clients become increasingly off-policy for the local model. The $\keep$ strategy alleviates this by retaining locally generated incorrect responses and by leaving the local response set unchanged on easier prompts where at least $\K/2$ responses are already correct.

\subsection{Experiments on higher heterogeneity}
In Table~\ref{tab:qwen3highet}, we compare our method against the baseline under a higher-heterogeneity client split. The corresponding bubble plot of the client-wise topic distributions is provided in the Appendix section \ref{app:expdetails}. While the local datasets in Table~\ref{tab:math_deepmath_results} were constructed by sampling client distributions from a Dirichlet distribution with $\alpha=0.3$, the split used in Table~\ref{tab:qwen3highet} corresponds to a more heterogeneous setting with $\alpha=0.1$. We generate equal-sized client datasets following the procedure of \citet{acar2021federated}. The $\keep$ method and a $\swap$ period of 2 is used here for response aggregation with $\dist$. The results show that, after 360 rounds of training, our proposed method outperforms FedAvg across different local-step configurations, consistent with the trends observed in the lower-heterogeneity setting.

\subsection{Ablation on $\swap$ period}
\begin{wraptable}{h}{0.5\textwidth}
\vspace{-0.5em}
\centering
\small
\begin{tabular}{@{}l|ccc@{}}
\toprule
Method & $\tau$=80 & $\tau$=90 & $\tau$=120 \\
\midrule
FedAvg-GRPO & 48.0 & 50.4 & 50.7 \\
FedProx-GRPO     & 46.5   & 48.0 & 47.7 \\
$\tau_\swap$ = 2    & 52.2 & 53.3 & \textbf{55.8} \\
$\tau_\swap$ = 4    & \textbf{54.3} & \textbf{54.3} & 52.5 \\
$\tau_\swap$ = 8    & 53.1 & 53.4 & 52.6 \\
\bottomrule
\end{tabular}
\caption{Pass@1 performance of Qwen3-1.7B across $\swap$ periods on DeepMath.}
\label{tab:qwen3_dp_ls}
\vspace{-1em}
\end{wraptable}

In Tables~\ref{tab:qwen3_dp_ls} and \ref{tab:qwen25math_dp_ls}, we study the effect of varying the $\dist$ period for the Qwen3-1.7B and Qwen2.5-Math-1.5B models. All experiments are run for 360 GRPO steps except for the $\tau$=80 column in Qwen3-1.7B, where the reported numbers correspond to the final accuracy after 320 steps.
Increasing the $\dist$ period reduces the frequency with which public data is incorporated into training, while decreasing it leads to more frequent public-data steps. Although more frequent public steps can in principle help limit drift through steps on shared prompts and response aggregation, they also increase the extent to which locally generated responses are replaced by globally correct responses, thereby introducing a larger off-policy shift especially in later stages of local training where the models drift further. Thus, smaller $\dist$ periods are not necessarily always beneficial. 

\begin{wraptable}{r}{0.4\textwidth}
\vspace{-0.8em}
\centering
\small
\begin{tabular}{@{}l|cc@{}}
\toprule
Method & $\tau$=90 & $\tau$=120 \\
\midrule
FedAvg-GRPO & 50.5 & 49.3 \\
FedProx-GRPO     & 48.0 & 49.4 \\
$\tau_\swap$ =  2       & 51.0 & 53.1 \\
$\tau_\swap$ = 4       & \textbf{52.3} & \textbf{53.6} \\
$\tau_\swap$ = 8       & 51.9 & 51.7 \\
\bottomrule
\end{tabular}
\caption{Pass@1 performance of Qwen2.5-Math-1.5B across different $\swap$ periods on DeepMath.}
\label{tab:qwen25math_dp_ls}
\vspace{-1em}
\end{wraptable}

While $\keep$ always induces higher reward variance within the rollout set used for model updates, which can often accelerate learning \citep{xu2025not,ye2025beyond}, this advantage may be attenuated in our setting. Unlike \citet{xu2025not}, where all rollouts are on-policy, our method injects responses that can be off-policy relative to the local model, and this mismatch can reduce the benefit of increased variance. Nevertheless, across settings and models and for $\dist$ periods of 2, 4, and 8, the proposed $\dist$ method with $\keep$ aggregation consistently outperforms the baselines.

\section{Conclusion}
\label{sec:conc}
In this work, we proposed PubSwap, a federated RLVR framework that combines LoRA-based local adaptation with coordination through shared public data. By interleaving private-data GRPO steps with public-data off-policy steps, our method improves cross-client alignment while preserving communication efficiency and privacy. Across mathematical and medical reasoning benchmarks, multiple model families, and varying heterogeneity levels, Fedavg-PubSwap consistently outperformed the baselines, with the consistent gains appearing in high-local-step regimes where communication efficiency matters most. Promising directions for future work include adaptive training strategies along multiple axes, rather than only adjusting the overall balance between public and private data. These include curriculum based strategies for sampling private data and public prompts, adaptive assignment of public prompts to clients based on informativeness or coordination value, and smarter response aggregation schemes that provide each client with the external responses most useful for its current policy. Stronger off-policy correction methods may further improve performance.

\section*{Acknowledgement}
This work was partially supported by the Office of Naval Research (ONR) under grant N000142412073 and by the U.S. National Science Foundation (NSF), including grant 2533813 to CJW. This work was also supported by NSF grants CCF-2045694, CNS-2112471, and CPS-2111751, as well as an AI2C Seed Grant. This work used Bridges-2 GPU resources at the Pittsburgh Supercomputing Center through allocation CIS260424 from the Advanced Cyberinfrastructure Coordination Ecosystem: Services \& Support (ACCESS) program, which is supported by NSF grants \#2138259, \#2138286, \#2138307, \#2137603, \#2312761, and \#2138296 \citep{access}.
\section*{Ethics Statement}
Our work focuses on communication-efficient federated training, which can support collaborative model development without requiring raw data sharing across organizations. We do not identify any unique ethical concerns beyond those already associated with federated learning and reasoning model deployment, though standard considerations around privacy, data governance, and responsible downstream use still apply.

\bibliography{colm2026_conference}
\bibliographystyle{colm2026_conference}

\appendix
\section*{Appendix}
\section*{LLM usage}
We used LLMs minimally, focusing on making sentences more concise to fit the page limit.

\section{Supporting Theoretical Derivations}
In this section, we analyze, in a simplified setting, the drift between two clients, $n$ and $n'$, induced by a single step of gradient descent on private data and by our 
$\keep$ method. This treatment is intended mainly to provide intuition for the behavior observed in the main text, rather than a fully general theorem-level statement. To keep the exposition compact, we do the analysis in a simplified setting and omit routine technical details:
\begin{itemize}
    \item The result is derived for SGD instead of AdamW and uses only one gradient update per GRPO step.
    \item We model heterogeneity and estimation noise assumptions directly at the objective-function level, rather than through data distributions and sampling (as is common in most federated learning analyses). This abstraction allows us to sidestep the technical complexities of the GRPO objective, including clipped importance ratios, gradient clipping induced bounded gradients, and KL regularization.
\end{itemize}

Analogous conclusions can be derived for the exact analytical expressions in GRPO with more extensive assumptions and bookkeeping.

\subsection{Understanding the Drift induced by Private Gradient Descent steps}
In this section, we quantify the drift between clients $n$ and $n'$ by deriving an upper bound on the effect of a single gradient descent step computed on a mini-batch of their respective private datasets. At the beginning of the communication round, all clients start with the same model $\theta_0$. Let $\Delta_t$ be the drift between the clients $n$ and $n'$ at local step $t$ defined as
\[
\Delta_t := \theta_n^{(t)}-\theta_{n'}^{(t)}.
\]
For the private local GRPO step, the parameters are updated as
\[
\theta_n^{(t+1)}=\theta_n^{(t)}+\eta \hat g_n^{(t)},
\qquad
\theta_{n'}^{(t+1)}=\theta_{n'}^{(t)}+\eta \hat g_{n'}^{(t)},
\]
where $\hat g_n^{(t)}$ is the stochastic GRPO gradient computed using the private minibatch $B_n^{(t)}$ at client $n$. Define the minibatch-conditioned expected private gradient
\[
g_B^{\priv}(\theta,\theta')
:=
\frac1b\sum_{x\in B}
\mathbb E_{\mathcal Y\sim \pi_{\theta'}(\cdot|x)^{\otimes K}}
\big[\nabla_\theta \ell_{\mathrm{GRPO}}(\theta;x,\mathcal Y)\big].
\]
For one gradient update per GRPO step $\theta = \theta'$, one can write the gradient term as
\[
g_B^{\priv}(\theta)
:=
\frac1b\sum_{x\in B}
\mathbb E_{\mathcal Y\sim \pi_\theta(\cdot|x)^{\otimes K}}
\big[\nabla_\theta \ell_{\mathrm{GRPO}}(\theta;x,\mathcal Y)\big].
\] 
where $\ell_{\mathrm{GRPO}}(\theta;x,\mathcal Y)$ denotes the GRPO loss at point $\theta$ with importance ratio also computed with the sampling policy $\theta$ on a rollout batch $\mathcal{Y}$ for the prompt $x$. The response sampling noise in $g_n^{(t)}$ can be quantified as follows
\[
\hat g_n^{(t)}=g_{B_n^{(t)}}^{\priv}(\theta_n^{(t)})+\xi_n^{(t)},
\]

Now,
\begin{equation}
    \Delta_{t+1}
=
\Delta_t+\eta\left(\hat g_n^{(t)}-\hat g_{n'}^{(t)}\right),
\label{eq:driftbasicdecomp}
\end{equation}
so
\[
\|\Delta_{t+1}\|
\le
\|\Delta_t\|+\eta\left\|\hat g_n^{(t)}-\hat g_{n'}^{(t)}\right\|.
\]

Adding and subtracting $g_{B_n^{(t)}}^{\priv}(\theta_{n'}^{(t)})$ gives
\begin{align*}
\hat g_n^{(t)}-\hat g_{n'}^{(t)}
&=
\left(g_{B_n^{(t)}}^{\priv}(\theta_n^{(t)})-g_{B_n^{(t)}}^{\priv}(\theta_{n'}^{(t)})\right) +
\left(g_{B_n^{(t)}}^{\priv}(\theta_{n'}^{(t)})-g_{B_{n'}^{(t)}}^{\priv}(\theta_{n'}^{(t)})\right) +
\left(\xi_n^{(t)}-\xi_{n'}^{(t)}\right).
\end{align*}
\newtheorem{assumption}{Assumption}

\begin{assumption}[Lipschitz private on-policy gradient]
There exists a constant $L_{\priv} > 0$ such that for any minibatch $B$ across clients of size $b$ and any $\theta, \theta' \in \Theta$,
\[
\left\|g_B^{\priv}(\theta) - g_B^{\priv}(\theta')\right\| \le L_{\priv}\|\theta - \theta'\|,
\]
where
\[
g_B^{\priv}(\theta) := \frac{1}{b}\sum_{x \in B} \mathbb{E}_{\mathcal{Y} \sim \pi_\theta(\cdot|x)^{\otimes K}}[\nabla_\theta \ell_{\mathrm{GRPO}}(\theta; x, \mathcal{Y})].
\]
\label{ass:liponp}
\end{assumption}
Since each local GRPO step performs a single gradient update from on-policy rollouts, this map jointly varies the sampling distribution and the optimization variable. This is stronger than requiring Lipschitzness in the optimization variable alone with a fixed sampling policy, which is sufficient for our case, as it additionally requires that the expected GRPO gradient varies smoothly as the response distribution shifts with the policy. Now using assumption \ref{ass:liponp} we have
\begin{align}
    \left\|g_B^{\priv}(\theta)-g_B^{\priv}(\theta')\right\|
\le
L_{\priv}\|\theta-\theta'\|.
\label{eq:liponp}
\end{align}

and using equations \ref{eq:liponp} and \ref{eq:driftbasicdecomp}
\[
\left\|\theta_n^{(t+1)}-\theta_{n'}^{(t+1)}\right\|
\le
(1+\eta L_{\priv})\left\|\theta_n^{(t)}-\theta_{n'}^{(t)}\right\|
+
\eta H_{n,n'}^{(t)}
+
\eta\left(\left\|\xi_n^{(t)}\right\|+\left\|\xi_{n'}^{(t)}\right\|\right),
\]
where
\[
H_{n,n'}^{(t)}
:=
\left\|
g_{B_n^{(t)}}^{\priv}(\theta_{n'}^{(t)})
-
g_{B_{n'}^{(t)}}^{\priv}(\theta_{n'}^{(t)})
\right\|.
\]

Further define the population private gradient to quantify the noise from minibatch selection
\[
g_n^{\priv}(\theta):=\mathbb E_{B_n}\left[g_{B_n}^{\priv}(\theta)\right] \quad \text{and} \quad H_{n,n'}^{(t)}
\le
\zeta_{n,n'}^{(t)}+\delta_n^{(t)}+\delta_{n'}^{(t)},
\]
where
\[
\zeta_{n,n'}^{(t)}
:=
\left\|g_n^{\priv}(\theta_{n'}^{(t)})-g_{n'}^{\priv}(\theta_{n'}^{(t)})\right\|,
\qquad
\delta_n^{(t)}
:=
\left\|g_{B_n^{(t)}}^{\priv}(\theta_{n'}^{(t)})-g_n^{\priv}(\theta_{n'}^{(t)})\right\|.
\]
Therefore,
\[
\left\|\theta_n^{(t+1)}-\theta_{n'}^{(t+1)}\right\|
\le
(1+\eta L_{priv})\left\|\theta_n^{(t)}-\theta_{n'}^{(t)}\right\|
+
\eta\zeta_{n,n'}^{(t)}
+
\eta(\delta_n^{(t)}+\delta_{n'}^{(t)})
+
\eta\left(\left\|\xi_n^{(t)}\right\|+\left\|\xi_{n'}^{(t)}\right\|\right).
\]
Note that the factor contributing to a large drift here will be $L_{\priv}$ this is taken as the maximum smoothness parameter over any mini batch \textit{across users} (Assumption \ref{ass:liponp}). Additionally, the term 
\[\zeta_{n,n'}^{(t)}
:=
\left\|g_n^{\priv}(\theta_{n'}^{(t)})-g_{n'}^{\priv}(\theta_{n'}^{(t)})\right\|,\]
is dictated by the dissimilarity of the avg gradient between the client $n$ and $n'$, and can be large in heterogeneous settings.
\subsection{Understanding the Drift induced by $\keep$ steps using public data}
Define the shared on-policy GRPO gradient at time $t$ as the expected gradient on the dataset $B_{\pub}^{(t)}$ chosen for the $\keep$ step at round $t$
\[
g_{\pub}^{(t)}(\theta)
:=
\frac1b\sum_{x\in B_{\pub}^{(t)}}
\mathbb E_{\mathcal Y\sim \pi_\theta(\cdot\mid x)^{\otimes K}}
\big[\nabla_\theta \ell_{\mathrm{GRPO}}(\theta;x,\mathcal Y)\big].
\]
Define the $\keep$ gradient at client $i$ on the public minibatch as
\[
g_{B_{\pub}^{(t)},\,i}^{\keep}\left(\theta;\Theta^{(t)}\right)
:=
\frac{1}{b}\sum_{x \in B_{\pub}^{(t)}}
\mathbb{E}_{\mathcal{Y} \sim Q_i^{\keep}(\cdot \mid x;\, \theta,\, \Theta^{(t)})}
\big[\nabla_\theta \ell_{\mathrm{GRPO}}(\theta; x, \mathcal{Y})\big],
\]
where $Q_i^{\keep}(\cdot \mid x;\, \theta,\, \Theta^{(t)})$ denotes the distribution
over response groups at client $i$ after applying the $\keep$ replacement rule:
client $i$ first generates $K$ responses from $\pi_\theta(\cdot \mid x)$, then replaces
up to $\bigl(\frac{K}{2} - c_i^{(t)}(x)\bigr)_+$ incorrect responses with correct
responses drawn from the globally collected pool
$\bigcup_{j \neq i} \{\text{responses from } \pi_{\theta_j^{(t)}}(\cdot \mid x)\}$.
The dependence on $\Theta^{(t)} := (\theta_1^{(t)}, \dots, \theta_N^{(t)})$ enters
through this donor pool. For each client $i$, define the $\keep$ distortion
\[
e_i^{(t)}\left(\theta, \Theta^{(t)}\right)
:=
g_{B_{\pub}^{(t)},\,i}^{\keep}\left(\theta;\Theta^{(t)}\right)
-
g_{\pub}^{(t)}(\theta).
\]
Then the stochastic $\keep$ gradient can be written as
\[
\hat g_i^{\keep,(t)}
=
g_{\pub}^{(t)}\left(\theta_i^{(t)}\right)
+
e_i^{(t)}\left(\theta_i^{(t)}, \Theta^{(t)}\right)
+
\xi_i^{\keep,(t)}.
\]
where $\xi_i^{\keep,(t)}$ is the noise term arising from randomness in response sampling. Hence
\begin{align*}
\hat g_n^{\keep,(t)}-\hat g_{n'}^{\keep,(t)}
&=
\Big(g_{\pub}^{(t)}\left(\theta_n^{(t)}\right)-g_{\pub}^{(t)}\left(\theta_{n'}^{(t)}\right)\Big)\\
&+
\Big(e_n^{(t)}\left(\theta_n^{(t)},\Theta^{(t)}\right)-e_{n'}^{(t)}\left(\theta_{n'}^{(t)}, \Theta^{(t)}\right)\Big) \\
\quad &+
\Big(\xi_n^{\keep,(t)}-\xi_{n'}^{\keep,(t)}\Big).
\end{align*}
Therefore,
\begin{align*}
\left\|\theta_n^{(t+1)}-\theta_{n'}^{(t+1)}\right\|
\le\;&
\left\|\theta_n^{(t)}-\theta_{n'}^{(t)}\right\| +
\eta\left\|g_{\pub}^{(t)}(\theta_n^{(t)})-g_{\pub}^{(t)}(\theta_{n'}^{(t)})\right\| \\
&+
\eta\left\|e_n^{(t)}\left(\theta_n^{(t)},\Theta^{(t)}\right)-e_{n'}^{(t)}\left(\theta_{n'}^{(t)},\Theta^{(t)}\right)\right\| \\
&+
\eta\left(\left\|\xi_n^{\keep,(t)}\right\|+\left\|\xi_{n'}^{\keep,(t)}\right\|\right).
\end{align*}
\begin{assumption}[Lipschitz public on-policy gradient]
\label{ass:lippub}
For any minibatch $B \subseteq D_{\pub}$ of size $b$, define
\[
g_B^{\pub}(\theta) := \frac{1}{b}\sum_{x \in B}
\mathbb{E}_{\mathcal{Y} \sim \pi_\theta(\cdot \mid x)^{\otimes K}}
\big[\nabla_\theta \ell_{\mathrm{GRPO}}(\theta; x, \mathcal{Y})\big].
\]
There exists a constant $L_{\pub} > 0$ such that for all such $B$ and all $\theta, \theta' \in \Theta$,
\[
\left\|g_B^{\pub}(\theta) - g_B^{\pub}(\theta')\right\| \le L_{\pub}\|\theta - \theta'\|.
\]
\end{assumption}
Note that at each round $t$, the server samples a minibatch $B_{\pub}^{(t)} \subseteq D_{\pub}$, and we write $g_{\pub}^{(t)}(\theta) := g_{B_{\pub}^{(t)}}^{\pub}(\theta)$ for brevity.

\begin{align*}
    \left\|\theta_n^{(t+1)}-\theta_{n'}^{(t+1)}\right\|
&\le
(1+\eta L_{\pub})\left\|\theta_n^{(t)}-\theta_{n'}^{(t)}\right\|
+
\eta\left\|e_n^{(t)}(\theta_n^{(t)},\Theta^{(t)})-e_{n'}^{(t)}(\theta_{n'}^{(t)},\Theta^{(t)})\right\|\\
&+
\eta\left(\left\|\xi_n^{\keep,(t)}\right\|+\left\|\xi_{n'}^{\keep,(t)}\right\|\right).
\end{align*}

Defining $\omega_i^{(t)}(\theta,\Theta^{(t)}):=\left\|e_i^{(t)}(\theta,\Theta^{(t)})\right\|$, we further get
\begin{align*}
    \left\|\theta_n^{(t+1)}-\theta_{n'}^{(t+1)}\right\|
&\le
(1+\eta L_{\pub})\left\|\theta_n^{(t)}-\theta_{n'}^{(t)}\right\|
+
\eta\omega_n^{(t)}\left(\theta_n^{(t)},\Theta^{(t)}\right)
+
\eta\omega_{n'}^{(t)}\left(\theta_{n'}^{(t)},\Theta^{(t)}\right)
\\
&+
\eta\big(\|\xi_n^{\keep,(t)}\|+\|\xi_{n'}^{\keep,(t)}\|\big).
\end{align*}

For the public data step with $\keep$ aggregation, recall
\[
e_i^{(t)}(\theta,\Theta^{(t)})
=
g_{B_{\pub}^{(t)},i}^{\keep}(\theta;\Theta^{(t)})
-
g_{\pub}^{(t)}(\theta).
\]
For a public prompt $x$, let $\mathcal Y_i^{(0)}(x)$ denote the fully local on-policy GRPO group for client $i$, and let $\mathcal Y_i^{(\keep)}(x)$ denote the final group used by $\keep$. Writing
\[
\mathcal G(\theta;x,\mathcal Y):=\nabla_\theta \ell_{\mathrm{GRPO}}(\theta;x,\mathcal Y),
\]
we have
\[
e_i^{(t)}\left(\theta,\Theta^{(t)}\right)
=
\frac1b\sum_{x\in B_{\pub}^{(t)}}
\left(
\mathbb E\left[\mathcal G(\theta;x,\mathcal Y_i^{(\keep)}(x))\right]
-
\mathbb E\left[\mathcal G(\theta;x,\mathcal Y_i^{(0)}(x))\right]
\right).
\]

Define the random variable number $M_i^{(t)}(x,\theta)$ denoting of replacements on prompt $x$ by $\keep$ at client $i$
\[
M_i^{(t)}(x,\theta)\leq\left(\frac K2-c_i^{(t)}(x)\right)_+,
\qquad
\alpha_i^{(t)}(x):=\frac{M_i^{(t)}(x,\theta)}{K}\leq \frac{1}{2}.
\]
Define the one-replacement GRPO distortion
\[
\beta_i^{(t)}\left(x;\theta\right)
:=
\sup_{(\mathcal Y,\widetilde{\mathcal Y})}
\left\|
\mathbb E\big[
\mathcal G(\theta;x,\widetilde{\mathcal Y})
-
\mathcal G(\theta;x,\mathcal Y)
\big]
\right\|,
\]
where the supremum is over all pairs $(\mathcal Y,\widetilde{\mathcal Y})$ that differ in exactly one coordinate, (happens when one local response in $\mathcal Y$ is replaced by a donor-correct response in $\widetilde{\mathcal Y}$), while the other $K-1$ responses are identical. If $M_i^{(t)}(x,\theta)=m$, let $\mathcal Y_i^{(0)}(x),\mathcal Y_i^{(1)}(x),\dots,\mathcal Y_i^{(m)}(x)$
be the sequence of intermediate groups obtained by performing the $m$ replacements one at a time, so that $\mathcal Y_i^{(m)}(x)=\mathcal Y_i^{(\keep)}(x)$. Then
\[
\mathcal G(\theta;x,\mathcal Y_i^{(m)}(x))
-
\mathcal G(\theta;x,\mathcal Y_i^{(0)}(x))
=
\sum_{s=1}^{m}
\Big(
\mathcal G(\theta;x,\mathcal Y_i^{(s)}(x))
-
\mathcal G(\theta;x,\mathcal Y_i^{(s-1)}(x))
\Big),
\]
and hence
\[
\left\|
\mathbb E[\mathcal G(\theta;x,\mathcal Y_i^{(\keep)}(x))]
-
\mathbb E[\mathcal G(\theta;x,\mathcal Y_i^{(0)}(x))]
\right\|
\le
M_i^{(t)}(x,\theta)\,\beta_i^{(t)}(x;\theta).
\]
Therefore,
\[
\omega_i^{(t)}(\theta)
\le
\frac1b\sum_{x\in B_{\pub}^{(t)}}
\mathbb E[M_i^{(t)}(x)]\,\beta_i^{(t)}(x;\theta)
=
K\cdot
\frac1b\sum_{x\in B_{\pub}^{(t)}}
\mathbb E[\alpha_i^{(t)}(x)]\,\beta_i^{(t)}(x;\theta).
\]

Hence, the additional drift induced by the $\keep$ step is controlled by the product of two factors: the fraction of responses that must be replaced, and the sensitivity of the GRPO gradient to a single such replacement.

This decomposition clarifies when the off-policy drift of $\keep$ is small. First, $\alpha_i^{(t)}(x)$ is small when the local model already performs reasonably well on the public prompt, so that few or no replacements are needed. This regime naturally becomes more common later in training or on easier public prompts. Second, $\beta_i^{(t)}(x;\theta)$ is small when donor-correct responses remain reasonably aligned with the local policy on the shared public prompt, so that replacing one response does not substantially perturb the GRPO update. In particular, $\beta_i^{(t)}(x;\theta)$ is expected to be smaller when clients have not drifted too far apart on the public data and when the GRPO group statistics are stable, since replacing a single response then only mildly changes the group-relative advantages - This partially formalizes the intuition that $\keep$ has more off policy tendency compared to FedAvg + GRPO.

Additionally, the drift now depends on $L_{\text{pub}}$ rather than $L_{\text{priv}}$. This can be a lot smaller, since $L_{\text{pub}}$ represents the maximum smoothness over mini-batches in the public dataset, whereas $L_{\text{priv}}$ is taken over all mini-batches across clients—a substantially larger and typically more heterogeneous set.

The above perspective also explains why $\keep$ is particularly appealing under strong client heterogeneity. For a purely private local step, the client-drift term is driven directly by the discrepancy between private gradients across heterogeneous clients. In contrast, under $\keep$, all clients are anchored to the same public minibatch, and the remaining discrepancy enters only through the off-policy distortion terms $\omega_i^{(t)}$. Therefore, whenever private-data heterogeneity is large while the public anchor remains effective that is, whenever  In words, heterogeneity create the regime in which public-data coordination is most useful.

\section{Additional Experimental Details and Results}
\label{app:expdetails}
\subsection{Client Dataset split}
For federated training, we partition both datasets across 4 clients using heterogeneous splits (based on topics); full details are provided in the appendix. In the MATH setting, each client receives 2,500 distinct non-overlapping samples, while the server maintains 1,250 public samples and a held-out test set of 1,250 samples. For DeepMath, we first randomly subsample 20K problems from the original 103K-problem corpus; each client then receives 4,000 samples, while the server again maintains 3,000 public samples and 1,000 held-out test samples. For the medical reasoning experiments, each client possesses 3956 datapoints while the size of the test set is 1977 and public dataset also has 1977 datapoints. 

Notably, in both settings, the public dataset is substantially smaller than the total amount of private data distributed across clients. The code required to reproduce the results presented in the paper is included in the supplementary material.

\begin{figure}[htbp]
    \centering
    \includegraphics[width=0.8\textwidth]{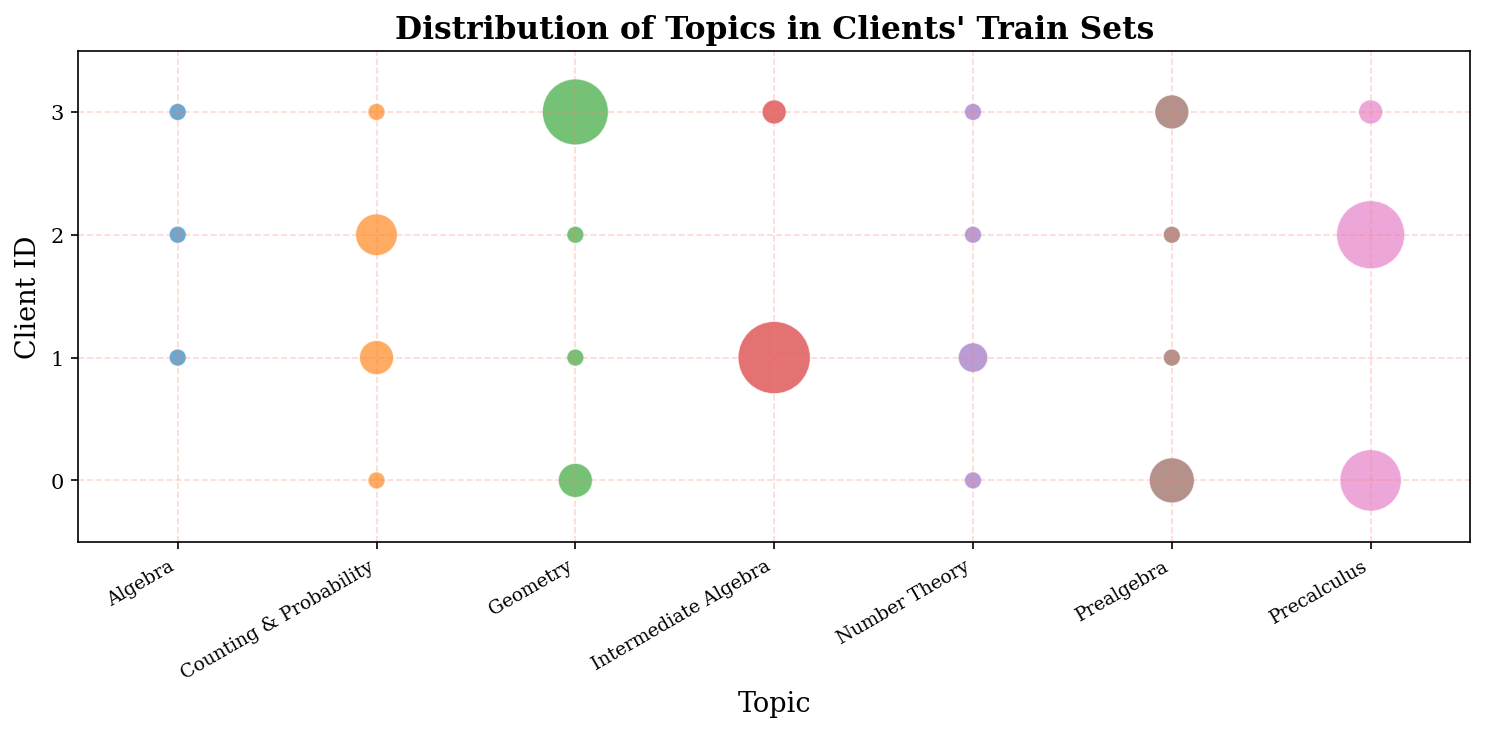}
    \caption{Bubble plot (quantized by 100 samples) of per client topic distribution heterogeneity for the MATH dataset experiments}
    \label{fig:my_figure}
\end{figure}

\begin{figure}[htbp]
    \centering
    \includegraphics[width=\textwidth]{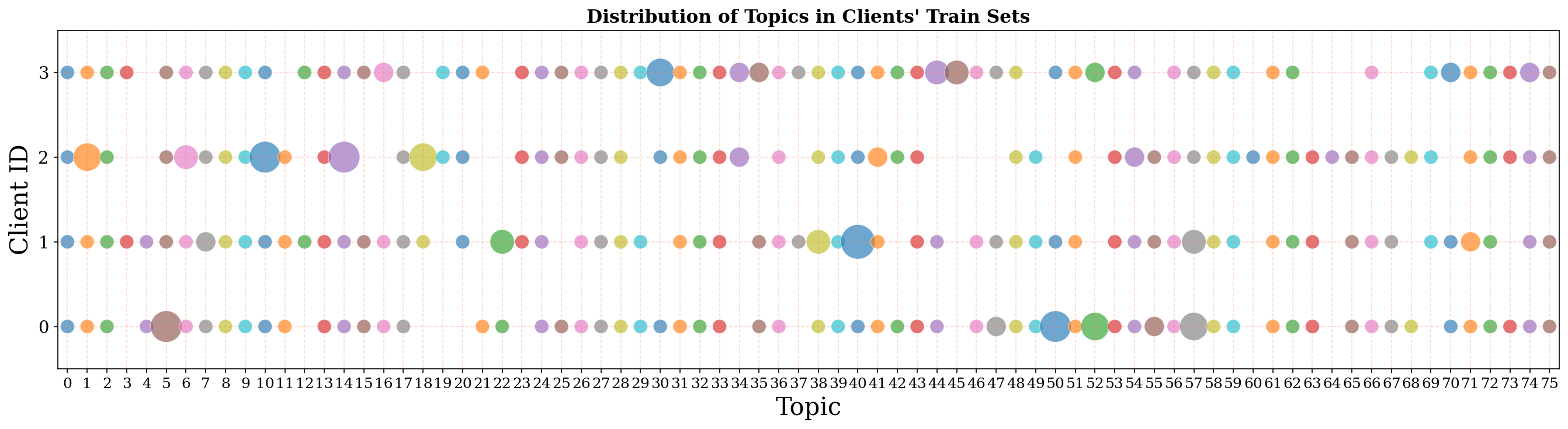}
    \caption{Bubble plot (quantized by 100 samples) of per client topic distribution (topic number) heterogeneity for the DeepMath dataset experiments}
    \label{fig:my_figure}
\end{figure}

\begin{figure}[htbp]
    \centering
    \includegraphics[width=\textwidth]{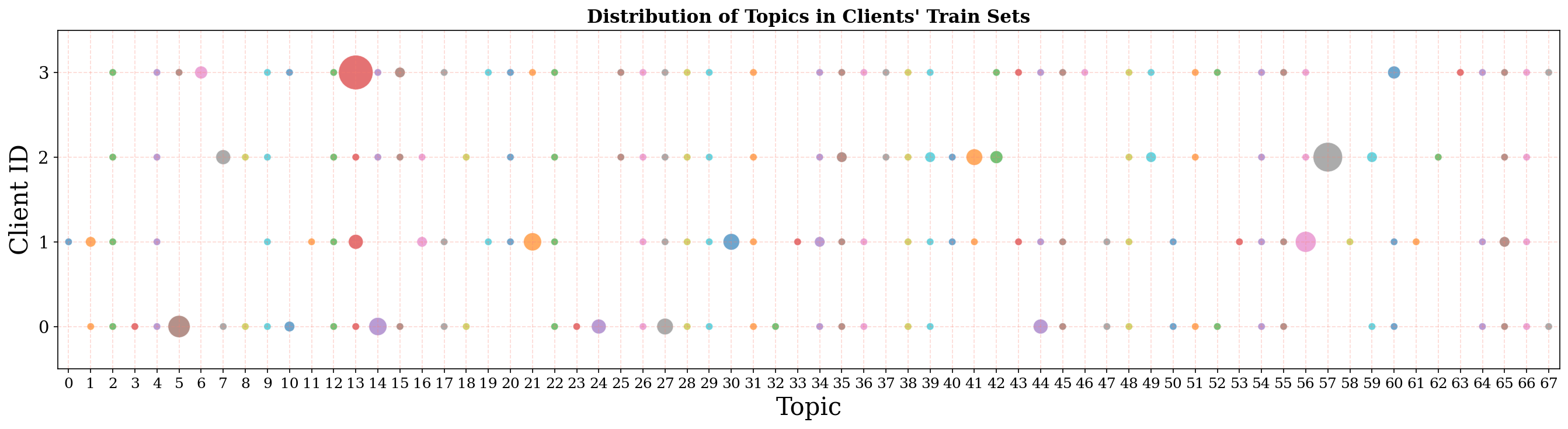}
    \caption{Bubble plot (quantized by 100 samples) of per client topic distribution heterogeneity for the high heterogeneity DeepMath dataset experiments (Table \ref{tab:qwen3highet})}
    \label{fig:my_figure}
\end{figure}

\subsection{Best checkpoint results}
In tables \ref{tab:math_deepmath_results_bcp}, \ref{tab:llama_med_bcp} we report the best checkpoint test performance of the models throughout the training rounds for our main experiments. The final checkpoint numbers are the same as the best checkpoint numbers in most high local step size regimes.
\begin{table}[h]
\centering
\small
\begin{tabular}{@{}ll|cccc|cccc@{}}
\toprule
\multirow{2}{*}{Model} & \multirow{2}{*}{Method}
& \multicolumn{4}{c|}{MATH}
& \multicolumn{4}{c}{DeepMath} \\
& & $\tau$=10 & $\tau$=40 & $\tau$=90 & $\tau$=120
& $\tau$=10 & $\tau$=40 & $\tau$=90 & $\tau$=120 \\
\midrule

\multirow{4}{*}{Qwen3-1.7B}
& Base model       & 55.2 & 55.2 & 55.2 & 55.2 & 14.9 & 14.9 & 14.9 & 14.9 \\
& FedAvg-GRPO  & \textbf{78.3} & 76.1 & 76.3 & 75.9 & 49.5 & 52.7 & 50.4 & 50.7 \\
& FedProx-GRPO      & 77.5 & 76.7 & 76.5 & 75.6 & 52.0 & 52.3 & 48.0 & 47.7 \\
& FedAvg-$\dist$      & 77.0 & \textbf{76.9} & \textbf{76.9} & \textbf{76.6} & \textbf{53.8} & \textbf{54.6} & \textbf{53.3} & \textbf{55.8} \\

\midrule

\multirow{4}{*}{\shortstack{Qwen2.5-\\Math-1.5B}}
& Base  model       & 58.4 & 58.4 & 58.4 & 58.4 & 34.9 & 34.9 & 34.9 & 34.9 \\
& FedAvg-GRPO  & \textbf{73.7} & 73.1 & 71.9 & 71.1 & 53.1 & 52.3 & 50.5 & 49.3 \\
& FedProx-GRPO      & 72.8 & 71.4 & 71.5 & 70.6 & 52.6 & 52.1 & 48.0 & 49.4 \\
& FedAvg-$\dist$      & 73.1 & \textbf{73.5} & \textbf{73.2} & \textbf{71.9} & \textbf{54.5} & \textbf{53.1} & \textbf{51.0} & \textbf{53.1} \\

\bottomrule
\end{tabular}
\caption{Best checkpoint pass@1 performance comparison on MATH \citep{hendrycks2021measuring} and DeepMath \citep{he2025deepmath} across different numbers of local steps ($\tau$). The $\keep$ method and a $\swap$ period of 2 is used here for response aggregation with $\dist$. }
\label{tab:math_deepmath_results_bcp}
\end{table}

\begin{table}[h]
\centering
\small
\begin{tabular}{@{}l|cccc@{}}
\toprule
Method & $\tau$=10 & $\tau$=40 & $\tau$=90 & $\tau$=120 \\
\midrule
Base model        & 49.2 & 49.2 & 49.2 & 49.2 \\
FedAvg-GRPO  & \textbf{59.7} & 58.9 & 57.9 & 57.7 \\
FedAvg-$\dist$      & 58.3 & \textbf{59.5} & \textbf{58.5} & \textbf{58.1} \\
\bottomrule
\end{tabular}
\caption{Best checkpoint pass@1 performance of the model Llama3.2-3B-Instruct on medical reasoning across different numbers of local steps ($\tau$). The $\keep$ method and a $\swap$ period of 2 is used here for response aggregation with $\dist$.}
\label{tab:llama_med_bcp}
\end{table}

\begin{table}[h]
\centering
\small
\setlength{\tabcolsep}{8pt}
\renewcommand{\arraystretch}{1.1}
\begin{tabular}{ll ll}
\toprule
\textbf{Parameter} & \textbf{Value} & \textbf{Parameter} & \textbf{Value} \\
\midrule
Pretrained Model                & Qwen3-1.7B      & Num Clients                & 4 \\
Grad epochs per GRPO step  & 2                    & Train Batch Size           & 8 \\
Max prompt length               & 1024                 & Max response length        & 2048 \\
LoRA rank                       & 32                   & LoRA alpha                 & 64 \\
LoRA modules                    & all linear           & Gen. per prompt     & 8 \\
Learning rate                   & $1 \times 10^{-5}$  & Clip ratio low             & 0.2 \\
Weight decay                    & 0.01                 & Gradient clip              & 1.0 \\
Clip ratio high                 & 0.25                 & KL loss coefficient        & 0.0001 \\
Entropy coefficient             & 0                    & Rollout engine             & vllm \\
Rollout temperature             & 0.7                  & Validation temperature     & 0.7 \\
Validation batch size           & 512                  & Remove padding             & Enabled \\
 Device                     & 2/4 $\times$ NVIDIA H100          & Aggregation method              & FedIT   \\
\bottomrule
\end{tabular}
\caption{Configuration for Qwen3-1.7B}
\label{tab:training_config}
\end{table}

\begin{table}[h]
\centering
\small
\setlength{\tabcolsep}{8pt}
\renewcommand{\arraystretch}{1.1}
\begin{tabular}{ll ll}
\toprule
\textbf{Parameter} & \textbf{Value} & \textbf{Parameter} & \textbf{Value} \\
\midrule
Pretrained Model                & Qwen2.5-Math-1.5B      & Num Clients                & 4 \\
Grad epochs per GRPO step  & 2                    & Train Batch Size           & 8 \\
Max prompt length               & 1024                 & Max response length        & 2048 \\
LoRA rank                       & 32                   & LoRA alpha                 & 64 \\
LoRA modules                    & all linear           & Gen. per prompt     & 8 \\
Learning rate                   & $1 \times 10^{-5}$  & Clip ratio low             & 0.2 \\
Weight decay                    & 0.01                 & Gradient clip              & 1.0 \\
Clip ratio high                 & 0.25                 & KL loss coefficient        & 0.0001 \\
Entropy coefficient             & 0                    & Rollout engine             & vllm \\
Rollout temperature             & 0.7                  & Validation temperature     & 0.7 \\
Validation batch size           & 512                  & Remove padding             & Enabled \\
 Device                     & 2/4 $\times$ NVIDIA H100          & Aggregation method              & FedIT   \\
\bottomrule
\end{tabular}
\caption{Configuration for Qwen2.5-Math-1.5B}
\label{tab:training_config}
\end{table}

\begin{table}[h]
\centering
\small
\setlength{\tabcolsep}{8pt}
\renewcommand{\arraystretch}{1.1}
\begin{tabular}{ll ll}
\toprule
\textbf{Parameter} & \textbf{Value} & \textbf{Parameter} & \textbf{Value} \\
\midrule
Pretrained Model                & Qwen3-4B-Instruct-2507     & Num Clients                & 4 \\
Grad epochs per GRPO step  & 2                    & Train Batch Size           & 4 \\
Max prompt length               & 1024                 & Max response length        & 2048 \\
LoRA rank                       & 32                   & LoRA alpha                 & 64 \\
LoRA modules                    & all linear           & Gen. per prompt     & 8 \\
Learning rate                   & $1 \times 10^{-5}$  & Clip ratio low             & 0.2 \\
Weight decay                    & 0.01                 & Gradient clip              & 1.0 \\
Clip ratio high                 & 0.25                 & KL loss coefficient        & 0.0001 \\
Entropy coefficient             & 0                    & Rollout engine             & vllm \\
Rollout temperature             & 0.7                  & Validation temperature     & 0.7 \\
Validation batch size           & 512                  & Remove padding             & Enabled \\
 Device                     & 2/4 $\times$ NVIDIA H100          & Aggregation method              & FedIT   \\
\bottomrule
\end{tabular}
\caption{Configuration for Qwen3-4B-Instruct}
\label{tab:training_config}
\end{table}

\begin{table}[h]
\centering
\small
\setlength{\tabcolsep}{8pt}
\renewcommand{\arraystretch}{1.1}
\begin{tabular}{ll ll}
\toprule
\textbf{Parameter} & \textbf{Value} & \textbf{Parameter} & \textbf{Value} \\
\midrule
Pretrained Model                & Llama-3.2-3B-Instruct & Num Clients                & 4 \\
Grad epochs per GRPO step  & 2                     & Train Batch Size           & 8 \\
Max prompt length               & 1024                  & Max response length        & 2048 \\
LoRA rank                       & 32                    & LoRA alpha                 & 64 \\
LoRA modules                    & \begin{tabular}[t]{@{}l@{}}q\_proj, k\_proj, v\_proj,\\ o\_proj, gate\_proj,\\ up\_proj, down\_proj\end{tabular}
                               & Gen. per prompt     & 8 \\
Learning rate                   & $1 \times 10^{-5}$   & Clip ratio low             & 0.2 \\
Weight decay                    & 0.01                  & Gradient clip              & 1.0 \\
Clip ratio high                 & 0.25                  & KL loss coefficient        & 0.0001 \\
Entropy coefficient             & 0                     & Rollout engine             & vllm \\
Rollout temperature             & 0.7                   & Validation temperature     & 0.7 \\
Validation batch size           & 512                   & Remove padding             & Enabled \\
Aggregation method              & FedIT                & Device                     & 2$\times$H100 \\
\bottomrule
\end{tabular}
\caption{Configuration for Llama-3.2-3B-Instruct}
\label{tab:training_config_llama}
\end{table}

\end{document}